# Three dimensional unique identifier based automated georeferencing and coregistration of point clouds in underground environment

*Sarvesh Kumar Singh[a], Bikram Pratap Banerjee[b], Simit Raval[a],\**

[a]*School of Minerals and Energy Resources Engineering, University of New South Wales, Sydney, NSW 2052, Australia;*

[b]*Agriculture Victoria, Grains Innovation Park, 110 Natimuk Road, Horsham, VIC 3400, Australia*

**Abstract:** Spatially and geometrically accurate laser scans are essential in modelling infrastructure for applications in civil, mining and transportation. Monitoring of underground or indoor environments such as mines or tunnels is challenging due to unavailability of a sensor positioning framework, complicated structurally symmetric layouts, repetitive features and occlusions. Current practices largely include a manual selection of discernable reference points for georeferencing and coregistration purpose. This study aims at overcoming these practical challenges in underground or indoor laser scanning. The developed approach involves automatically and uniquely identifiable three dimensional unique identifiers (3DUIDs) in laser scans, and a 3D registration (3DReG) workflow. Field testing of the method in an underground tunnel has been found accurate, effective and efficient. Additionally, a method for automatically extracting roadway tunnel profile has been exhibited. The developed 3DUID can be used in roadway profile extraction, guided automation, sensor calibration, reference targets for routine survey and deformation monitoring.

**Keywords:** LiDAR scanning; 3D point cloud; Scan matching; Control points; 3DUID; 3DReG

## 1. Introduction

Point clouds obtained from terrestrial or mobile laser scanning plays a critical role in infrastructure mapping and development projects, which is well suited for civil, mining and transportation industries. In an indoor or underground environment, direct georeferencing of the 3D point cloud is challenging due to the absence of a global navigation satellite system (GNSS) signal. The predominant practice under such a situation is to transfer datum from a GNSS aided environment to GNSS denied environment using conventional survey method. In the process, few easily distinguishable control points are marked using a total station to obtain global three-dimensional (3D) coordinates. Correspondence between local and global coordinate system is then established by recognising the control points in the point cloud [1,2]. The process involves manual identification of installed control points from the scanned point cloud, which is often time-consuming and potentially erroneous. Other studies have partially automated the process by georeferencing point clouds using coregistration of subsequent laser frames by matching installed control points [3,4]. The accuracy of georeferencing usually depends on how correctly the control points are identified in the lidar data [5,6]. The accuracy in georeferencing has an irrefutable impact on localisation and change detection capability in multi-temporal point cloud data. In the absence of ground control targets (GCTs), multi-temporal datasets are coregistered using algorithms such as iterative close point (ICP) and normal distribution transform (NDT) to obtain mutually aligned point clouds [7]. However, coregistration depends on several factors essential for cloud matching, such as the presence of discriminative features, surface reflectivity, structure repeatability and geometrical symmetry in the environment. In mobile laser scanning, environments with symmetry and repeatability such as an underground mine, tunnel or large indoor space leads to aliasing issues, i.e., false laser scan matches in coregistration due to the absence of the adequate number of distinct features. The aliasing issues results in inaccurate localisation, mapping drift and distorted objects in spatio-temporal point cloud data [8]. To improve coregistration, researchers have progressively developed techniques using point descriptors, which are generated for individual points in the point cloud to uniquely highlight characteristics such as curvature, local point geometry and surface variation in the environment [9]. Some of the widely used descriptors include eigenvalue descriptor [10,11], fast point feature histogram [12], radial surface descriptors [13], height gradient histogram [14] and scale-invariant feature transform [15]. Point descriptors though mostly effective are computationally intensive to generate and can produce unreliable results for environments with a low number of distinguishable features. Consequently, the generation of descriptors is avoided when real-time or rapid solutions are required. Alternatively, in a featureless and GNSS denied environment, additional information is acquired either through multi-sensor integration or by introducing artificially induced discernible active or passive GCTs. The use of discernible GCTs is a widely used simple and effective approach. The GCTs are of mainly two categories – (1) active and (2) passive.

Active GCTs require power to operate or are stimulated in presence of external active power source. Existing research based on the active tags sensing primarily include radio frequency identification (RFID) tags, or a wireless sensor network [16,17]. RFID tags together with RFID reader are used to map the drift in the underground mine [18,19]. Coordinates of the installed location of tags are stored in the internal memory of the chipsets along with its unique identifier (ID), which is used in registration and coregistration. In coregistration, the process involves the initial alignment of multiple point clouds using tag's coordinates, then a fine refinement is performed using the iterative closest point (ICP) algorithm. When using simultaneous localisation and mapping (SLAM) to scan a large underground space, the inertial measurement unit (IMU) sensors tend to accumulate error over time that results in an inaccurate 3D map. Previous research indicates that the installation of RFID tags can be useful along long sections of the underground tunnels to reduce data drift [18]. Combining data from RFIDs, odometer and IMU has been found effective to reduce IMU drift incurred for localisation and mapping [20]. A few studies have targeted to avoid using inertial sensors altogether and focused solely on using a laser scanner and RFIDs for accurate localisation and mapping [18,21,22]. Inconsistencies in laser frame stitching were observed as the distances between the RFID tags were increased. Although active

GCTs are accurate, they need the power to operate. For operational scenarios posing strict intrinsic safety requirements, statutory regulations or fire related hazards, such as an active underground coal mine, the use of active sensors is often a constrain. Under such scenarios, passive GCTs provide a sensible alternative tool for 3D mapping.

Passive GCTs do not require power and can be recognised through optical cameras or laser scanners using the spectral or geometrical characteristics of the target, respectively. GCT with retro-reflective materials exhibit higher intensity than the background structures which in turn aids/helps the process of georeferencing and coregistration of the point clouds [23–25]. An approach based on intensity relies on the reflectivity of retroreflective GCTs, which is likely to attenuate over time due to corrosion and dust. Additionally, intensity based approach is only suitable if the laser scanner is equipped to capture the intensity information. Therefore, geometric GCTs such as panels and spheres are often preferred over reflective targets for georeferencing and coregistration [26–28]. The recognition of geometric centres for spherical GCTs is achieved using surface reconstruction and tagged with an associated coordinate, the method is effective even in a low-resolution point cloud. However, identical GCTs which are symmetrical in structure are virtually indistinguishable, therefore supervised labelling of GCTs is necessary based on field notes detailing the local environment of the installed location. The technique is often tedious, error-prone and unsuitable at large scale for automated georeferencing and data coregistration applications. The necessity of unique identifiers as GCTs led to the development of a 3D barcode that was shown to reduce lateral and axial error in localisation of a vehicle [29]. However, the showcased 3D barcode was rarely used in future research due to its arduous construction and recognition challenges in the noisy and sparse point cloud. As a result, a complementary optical camera sensor was introduced for recognising 2D barcode/QR code patterns instead of 3D for localisation [30–32]. The outcome of an optical camera-based approach depends on the lighting condition and GCT recognition is often compromised in sub-optimally lit underground or indoor space. Further, the use of complementary sensors increases system cost. Therefore, GCTs must be recognised directly from the 3D point cloud without relying on additional sensors for recognition while keeping the identity of each target intact. An approach involving 3D geometric structures placed opposite to each other on the mine wall was presented for accurate localisation and mapping using triangulation, where the geometric targets were identical, but the lengths of edges were varied to attach a unique identity [33]. The georeferencing of the point cloud was achieved by surveying targets using a total station and then recognising the targets in the point cloud to establish correspondence. Most of the passive target based approaches use an optical camera to detect a 2D pattern or rely on the intensity of the retro-reflective target in case of laser scanning [23,30,31]. A major limitation of such a system is that it fails in a dark environment and is often affected by dust and dirt. The other GCTs which can be recognised in 3D point clouds are either difficult to construct or hard to decode.

Monitoring of underground space, particularly coal mine, is still a challenge due to major challenges such as suboptimal lighting, fire related hazards, regulatory requirements on the use of active GCT's due to intrinsic safety and lack of global frame of reference. This study develops and proposes a three dimensional unique identifier, referred to as 3DUID, to solve georeferencing and data coregistration in sensitive underground space for automated monitoring applications. The developed 3DUID is passive in nature, easy to construct, easily decodable, low-cost, has attributes for ease in surveying, and is robust against minor noise which is typical in point clouds for recognition purposes. Moreover, complementary optical sensors are not needed, and the recognition is accomplished directly using 3D point cloud. A robust workflow, called three dimensional unique identifier based registration and georeferencing (3DReG), is developed which extracts and decodes 3DUID and automatically georeference the scanned point cloud or mutually coregisters multiple point clouds. The experiment was conducted in an operational underground coal mine, which had mapping challenges including repetitive sections, sub-optimal lighting and low material reflectivity. Results demonstrate the efficacy of 3DReG in obtaining accurate georeferencing and data coregistration. Further, an automatic extraction of roadway profile, mainly horizontal and vertical clearance, has also been demonstrated which was compared against the ground truth data for validation. The aim of the study is not to improve the simultaneous localisation and mapping algorithm (SLAM) but to develop a spatial referencing framework for accurate and automated georeferencing and coregistration of point clouds.

## 2. Materials and Methods

### 2.1 3DUID development

An investigation on the mapping characteristics of the laser scanner was deemed important to identify the suitable shape, dimension and material for 3DUID development and recognition. Three laboratory tests were conducted to identify the dimensions and the suitability of a material required for construction (Fig. 1). In the first laboratory test, several patterns in the form of a barcode carved on a polystyrene board and ∧ shaped structure were used. Rectangular blocks with spacing varying from 1 mm to 15 mm (the top row in Fig. 1a) as well as the spacing of 30 mm, 40 mm and 50 mm (the second row in Fig. 1a) were created. The ∧ shaped pattern with a depth of 30 mm and spacing ranging from 1 mm to 60 mm was also created (the third row in Fig. 1a). The test was designed to aid in determining the separation threshold needed between individual elements or block of 3DUID for effective pattern recognition when scanned from a distance. For the selected corridor type environment, a mine tunnel, the distance between the opposite walls is within the range of 5 m. As the laser equipment scans the tunnel from within the tunnel, the maximal possible distance between the laser and 3DUID at the moment of the closest pass is always < 5 m. Therefore, the scanning distance between the sensor and potential targets during the testing and development of 3DUID was kept at 5 m. In the generated point cloud, the structures with separation less than 2.2 cm were not observed and the ∧ shaped pattern showed that the minimum measurement required for discernible separation was 3 cm (Fig. 1d).

A second laboratory test was conducted to observe the effect of material reflectivity on the separation threshold. Two ∧ shaped patterns, coated with moderately and highly reflective materials, were installed and scanned from a distance of 5 m (Fig. 1b). Reflective materials usually provide better laser backscatter, aiding in increasing the number of scanned points, which in turn enable better characterisation and segregation of scanned object. The separation observed in ∧ shaped pattern was 2.9 cm and 5.2 cm for moderately reflective and highly reflective material, respectively (Fig. 1e).

To verify the results from first and second tests, a third laboratory test was undertaken where several square blocks with dimensions ranging from 5 mm to 45 mm were installed at a separation of 3 cm. The three columns of blocks consisted of polystyrene panels covered with a moderately reflective, highly reflective and very high reflective material (Fig. 1c). The square blocks with a dimension greater than 3 cm were identified with confidence although in the case of high and very high reflective material the incurred noise made it hard to segregate the square blocks (Fig. 1f). Interestingly, it was found that using a high reflective material for recognition is not ideal because of incurred noise caused by multipath of laser rays between spaced adjacent blocks due to specular reflection. The multipath leads to false range measurement as the signal return is misdirected and reception time is prolonged leading to distorted geometry of the target.

The overall results indicated a block size of 3 cm (approx.) as the minimum detectable linear unit using point cloud scans from a distance of 5 m. To account for uncertainties in physical deployment and laser scanning of 3DUID inside a mining tunnel with dusty conditions, the dimension of elemental blocks was fixed to 6 cm (double the minimum mappable unit) for efficient recognition.

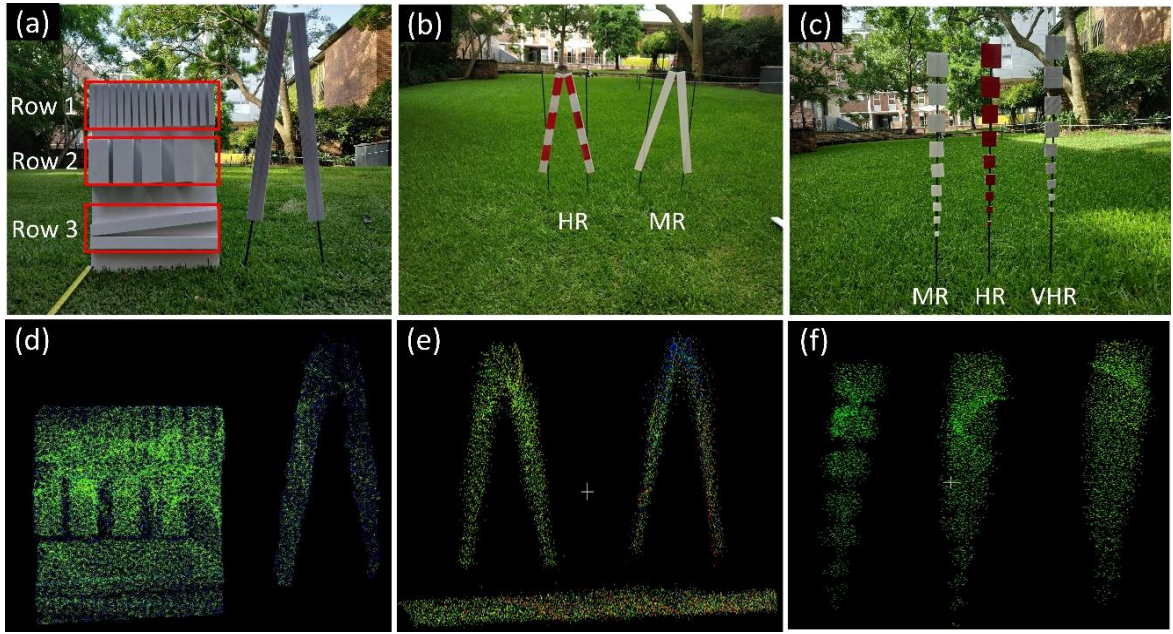

Fig. 1. Experimental tests for prototype development: (a) A barcode pattern with different length and spacing. Row 1 has spacing ranging from 1 mm to 15 mm, row 2 has spacing 30, 40 and 50 mm, and row 3 is a ∧ shaped pattern. (b) A ∧ shaped pattern with highly reflective (HR) and moderately reflective material (MR), and (c) a square block with 2.5 cm spacing made of moderate (MR), high (HR) and very high reflective (VHR) material was placed in a well-lit environment. (d) Point cloud corresponding to figure (a), (e) point cloud corresponding to figure (b), and (f) point cloud corresponding to figure (c).

Based on the shape, dimensions and material obtained through experiments, the size of the 3DUID pattern was computed. To generate several unique 3DUID tags, a simplistic algorithm shown in algorithm 1 was developed. A window size of $5 \times 5$ was selected for generating unique pattern, where each grid of the window corresponds to either 0 (absent or missing block space) or 1 (present block space) as an index value. Since a total of 25 blocks were available, index values for each block were systematically generated using computational combinations from $^{25}C_1$ to $^{25}C_{25}$. A boundary frame was attached to the generated pattern by adding 1's at the edges of the $5 \times 5$ pattern (Fig. 2). While generating a unique pattern, the algorithm checks for 'hanging pieces', i.e., the blocks of 1's not connected to any edge (Fig. 2a). It is physically impossible to create these virtual algorithmic constructs with hanging pieces in the 3DUIDs, therefore, patterns with hanging pieces were filtered using a connected component algorithm (Algorithm 1, line 8). A conceptual view of the prototype pattern is shown in Fig. 2b. With a window size of $5 \times 5$ and removal of hanging pieces, the 3DUID generator algorithm was able to simulate over 23.7 million unique patterns of potential 3DUID. The window size can be varied depending on the requirements of a given site. For instance, the size of window can be lowered to $3 \times 3$ to generate 496 unique patterns and save on material cost in construction of 3DUIDs. The red arrowed lines in Fig. 2b represent dimensions obtained after experimental tests. During the construction of 3DUID, '0' shown in black results in a void in the 3DUID, i.e., the material has been removed at those locations from a solid panel. The voids are aimed at decoding the pattern through 3D depth perception in the point cloud (Fig. 2c).

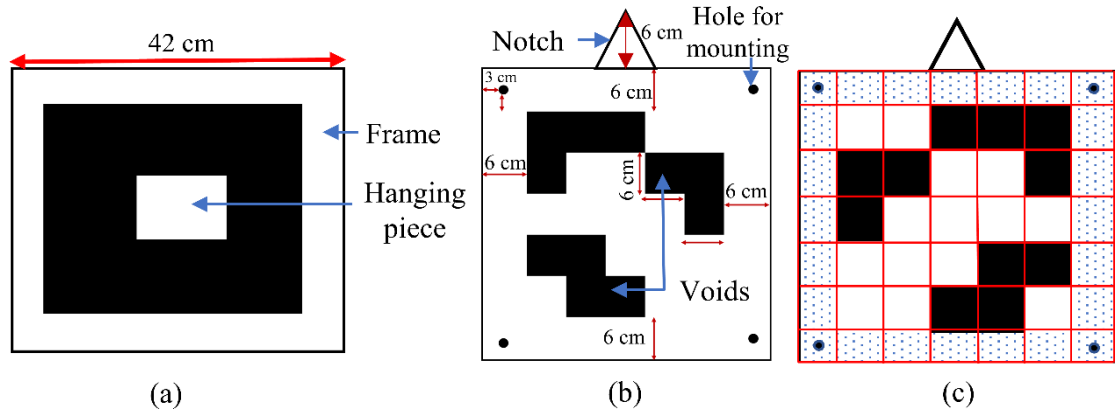

Fig. 2. (a) Boundary frame of the 3DUID and the hanging piece, which is to be avoided during construction, and (b) a conceptual 3DUID prototype with red lines indicating the dimension obtained using a minimum mappable unit, and a triangular notch on the top for automated georeferencing and coregistration. The black blocks represent voids where the material has been removed. (c) Gridding approach for coding and decoding 3DUID.

| **Algorithm 1**: 3DUID pattern generation |
|---|
| **Input:** Window size ($m \times m$) |
| **Output:** All the possible 3DUID patterns |
| 1: Generate a grid of $m + 2 \times m + 2$, where the factor '2' is grid associated with boundary frame |
| 2: Store 1 in the boundary grids, i.e. pad first row, last row, first column and last column with 1 |
| 3:   **for** i = 1 to $m \times m$ |
| 4:      index = $^{m \times m}C_i$ |
| 5:      Store 1 at given index location in $m \times m$ pattern and 0 elsewhere |
| 7:      Check for hanging pieces using connected component (pattern of 1 not connect to edge in $m + 2 \times m + 2$ pattern) |
| 8:      **if**   number of connected component > 1 |
| 9:           ignore pattern and continue loop |
| 10:     **else** |
| 11:          store 3DUID pattern |
| 12:     **end if** |
| 13:   **end for** |
| 14: Scale pattern or modify dots per inch (dpi) to convert the pattern to world dimensions for 3D printing or laser cutting (optional) |

A 1.5 cm thick opaque white acrylic panel was used for 3DUID construction and the selected pattern was carved by laser cutting for the voids. Since the obtained minimum mappable unit was 3 cm from a scanning distance of 5 m, the size of each grid in 5 × 5 pattern was kept 6 cm account for uncertainties at an operational mine site (Fig. 3a). Thus, the panel had a dimension of 42 cm × 42 cm including extra grids on the boundary for structural support (Fig. 3a). An equilateral triangular notch of 6 cm was added at the apex for ease in surveying, automated georeferencing and coregistration process. Holes were drilled on the four corners to support the mounting mechanism. A prototype of the generated 3DUID patterns is shown in Fig. 3b.

## 2.2 Study area, 3DUID installation and laser scanning

The study area was a section of an underground coal mine located in Southern Coalfields, New South Wales, Australia. The study area, approximately 850 m long, was selected to test the 3DUID in a range of challenging conditions such as unavailability of GNSS signals, intrinsic safety/fire related hazards, limited to no lighting, repetitive geometry, lack of discriminative features, low surface reflectivity and dusty environment. The experimental conditions were ideal for testing the efficacy of 3DUID assisted and unassisted scan registrations. A total of thirteen unique 3DUID patterns were generated, constructed and installed against the wall at an offset distance of 10 - 15 cm to enable depth perception for pattern recognition at the mine site (Fig. 3c). Each 3DUID tag was installed over the underground wired mesh support using tie-on metal wires hooked at four corners (Fig. 3d). A laser scanner mounted on a mining vehicle performs an ordinary pass-by scan of the tunnel in areas with the installed 3DUID (Fig. 3e). The accuracy of georeferencing and data coregistration depends on the number and distribution of 3DUID tags across the section of the tunnel. Increasing the number of installed 3DUIDs as GCTs is expected to increase both accuracy and cost. Therefore, a variably spaced strategic installation of 3DUIDs was adopted in this study to identify the optimal spacing in terms of both accuracy and cost effectiveness (Fig. 3f). Four discrete spacing levels were defined, i.e., 25 m, 50 m, 100 m and 200 m, such that for each spacing level a total of five 3DUID tags were available, this also facilitated leave-one-out error analysis. For georeferencing process, a surveying exercise was undertaken where the notch on the top of each 3DUID tag was surveyed with a total station (Trimble M3, California, USA). The coordinate system used was the map grid of Australia 1994 (MGA94) which is based on the geocentric datum of Australia 1994 (GDA) and the Australian height datum (AHD).

A SLAM based mobile laser scanner (ZebRevo, GeoSlam, Denver, USA) was used to acquire point cloud data in the test area. The scanner has a measurable range of 30 m with a laser pulse of 905 nm wavelength, classified as a reasonably safe to use class 1 laser product. The scan rate of the laser is around 43,200 points/sec with a range error of ±3 cm [34]. The scanner was mounted on a mining vehicle which was driven at an average speed of 10 km/hr. The vehicle speed was kept at the most practical low pace to avoid IMU error induced mapping drift that might be incorporated due to sudden jerks or low point density in individual laser frames. Furthermore, the vehicle was slowed down to 5 km/hr at the 3DUID location to capture more points or finer details on 3DUID for improved recognition. Two datasets were collected on the same day while avoiding any environmental change to evaluate georeferencing and coregistration processes. The collected point clouds of two scans were post-processed using the SLAM algorithm provided by GeoSLAM [34].

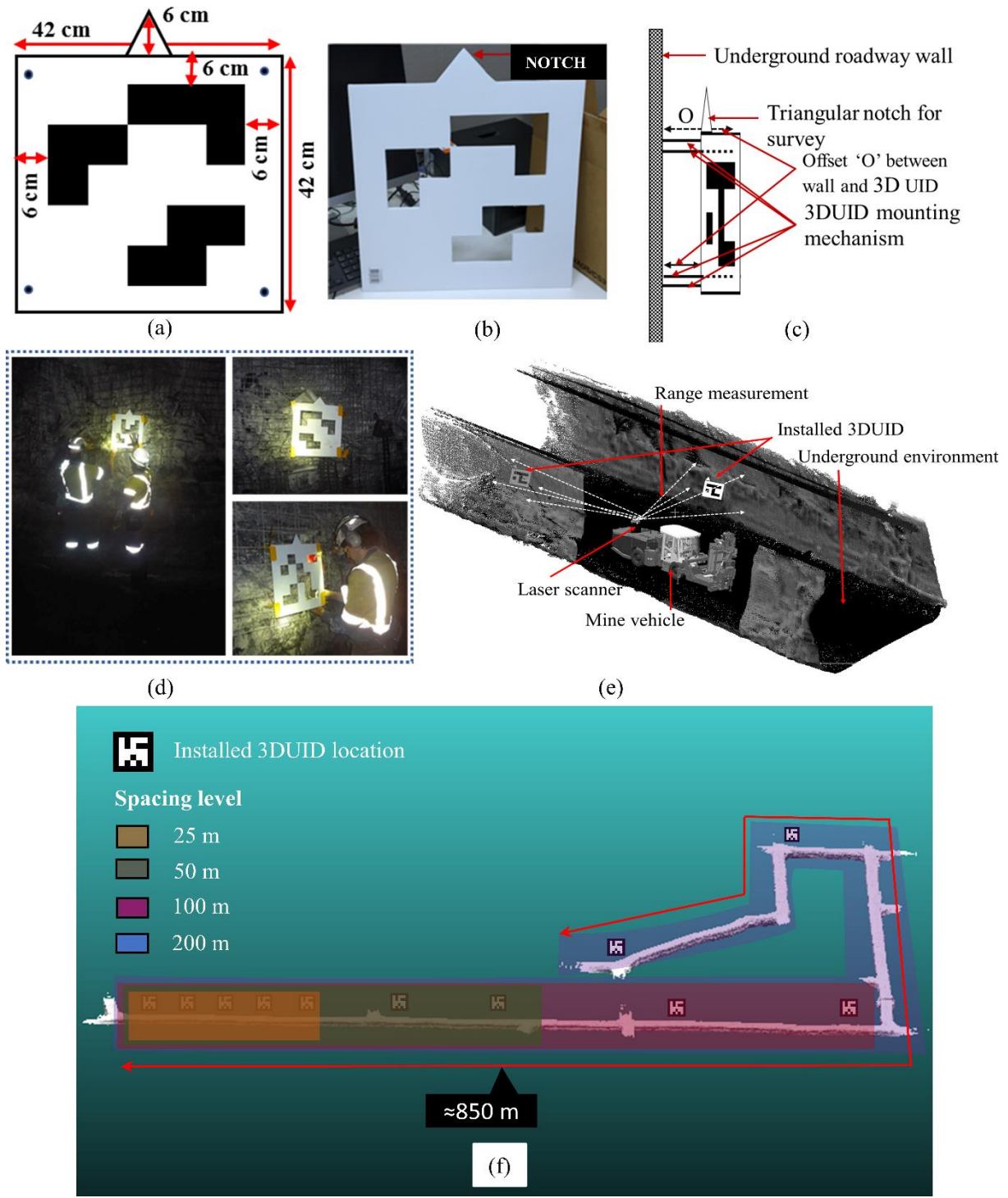

Fig. 3. (a) Selected dimensions for 3DUID construction shown with an example, (b) developed prototype with a triangular notch on top, (c) schematic view of 3DUID installation at an offset distance from the wall for depth perception, (d) actual installation of 3DUID at the mine site, (e) conceptual scanning of installed 3DUID, and (f) division of study area into several subsections to investigate the impact of inter 3DUID spacing on georeferencing and data coregistration.

## 2.3 Methodology for 3DReG

A robust workflow was developed to identify 3DUID in point cloud data (Fig. 4). The SLAM processed raw point cloud was first filtered to remove noise and outliers, and then an algorithmic data segmentation was done to reduce processing time for 3DUID identification. In the segmented point cloud, a 3DUID recognition algorithm was executed for automated georeferencing and data coregistration. The detailed explanation of workflow is presented in Section 2.3.1 through to Section 2.3.4.

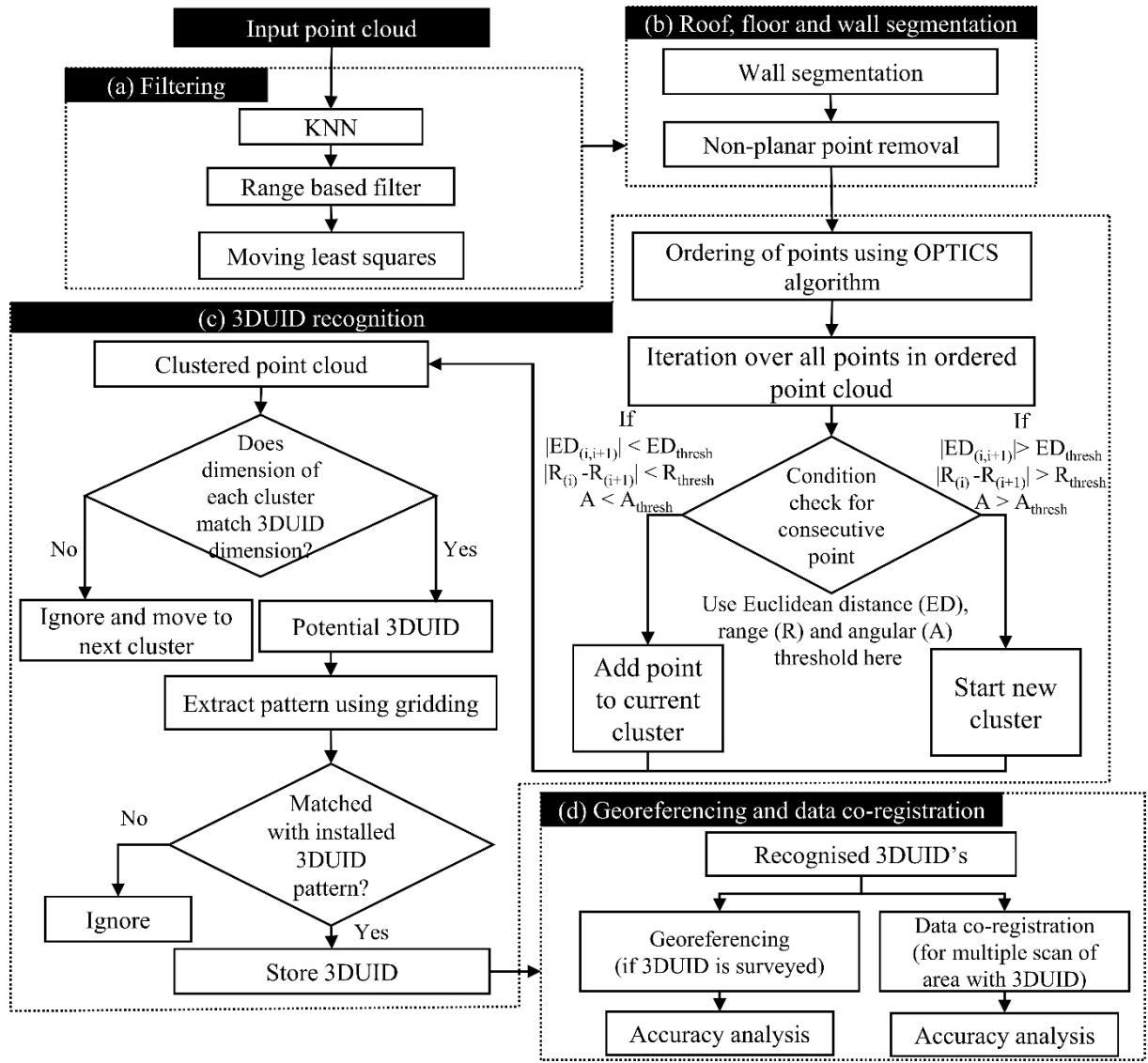

Fig. 4. Stages of 3DReG workflow to facilitate automated georeferencing and data coregistration. (a) Filtering of point cloud data to remove erroneous and uncertain points, (b) segmentation of roof, floor and wall to reduce the processing time for 3DUID recognition, (c) processing workflow for 3DUID recognition using ordered point cloud obtained from OPTICS algorithm, and (d) georeferencing using surveyed 3DUID and multiple point cloud coregistration through 3DUID correspondence.

### 2.3.1 Filtering

The filtering step is essential to remove erroneous or noisy points in the dataset which might have been captured due to false range and IMU measurements. The factors leading to range and IMU error include specular reflection, beam divergence, multipath, sensor perturbation, sudden jerks and low reflectivity of the surface. The presence of noise in the dataset has a considerable impact on accuracy and hence are required to be removed. Three filters K-nearest neighbour (KNN), range filter and moving least squares filter were applied in sequence to remove noise in the dataset. KNN filter is one of the most common point cloud filtering technique which removes selective outliers based on the distribution of 'k' nearest points around a query point [35]. If the distance of query point is greater than mean distance plus certain times the standard deviation (where k=10 and σ =1), in this study), then the point is flagged as noise. The point cloud also contained a high number of points with range based uncertainties which were removed using a range limiting filter. Since the spatial error in a point depends on the maximum measurable range of the scanner, therefore, the points which are scanned from a farther distance tends to incorporate more error due to sensor beam divergence [36,37]. Consequently, the scanned point cloud remains less effective in mapping fine level structural details, which also impacts the recognition of 3DUID and subsequent georeferencing accuracy. Additionally, such points cause unevenness in the surface leading to high variation in surface normal vectors and curvatures affecting the data coregistration. For range limiting filter, in a small spherical region, the range values of points were adjudged and points with a range greater than mean range plus one standard deviation were removed. Finally, a moving least square filter was applied to project the remaining erroneous points towards a best fit first-order polynomial surface, until they were within the specified threshold of mean distance plus one standard deviation [38,39]. For each of the three filters, the radius of the spherical region of interest to compute standard deviation at a query point was set to 0.04 m, which is neither too big nor too small to vary the geometrical characteristic of the scanned tunnel surfaces.

### 2.3.2 Roof, floor and wall segmentation

Recognition of 3DUID tags, installed on tunnel walls, was done by masking irrelevant roof and ground points to reduce computation time. The laser scanners trajectory was used in the process to measure the angle projected at a point between the nearest sensor location and the XY plane (Fig. 5a), using Eq. 1.

$$Angle = \left| tan^{-1} (Z_p - Z_s) \Big/ \left( (X_p - X_s)^2 + (Y_p - Y_s)^2 \right)^{0.5} \right| \quad (1)$$

where, *p* represents a point, *s* represents the nearest sensor location, and *X*, *Y*, *Z* are three axial coordinates.

If the projected angle at a point was greater than 30° or less than -30° then the point belonged to the roof or floor otherwise it belonged to the wall. After all the points were checked using the angle threshold, the two walls, roof and floor gets segmented (Fig. 5b). The two walls, and the floor and roof were separated using a connected component filter [40] and was given a false colour for better visualisation. The point cloud of two walls was used in further processing for 3DUID recognition.

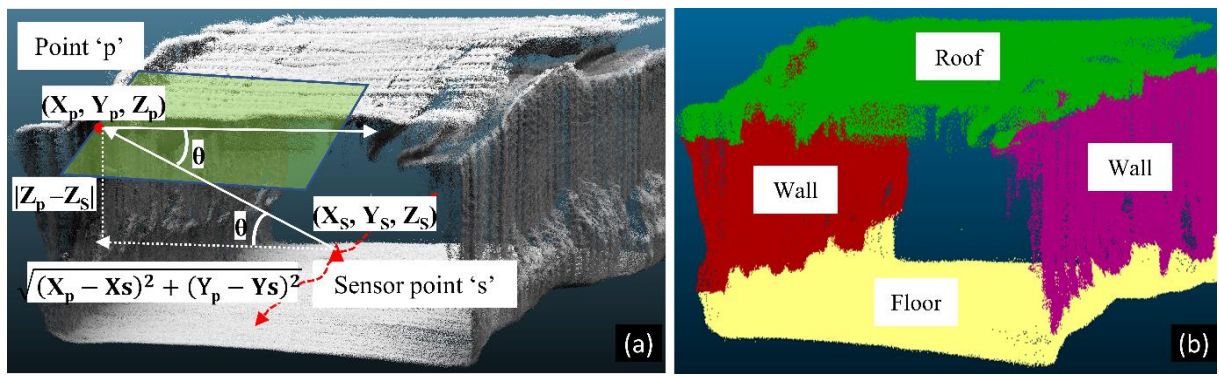

Fig. 5. (a) Angle projected at a point between the nearest sensor location and the XY plane (shown with green shaded area). Whenever this angle was less than the specified threshold value, the point was stored. The process quickly segments the wall which may contain potential 3DUID. (b)The segmented roof, floor and walls shown with different colours for better visualisation.

### 2.3.3 3DUID recognition and decoding

The point cloud of the wall was used in processing for 3DUID recognition. As a final filtering step to effectively recognise 3DUID, all the points which were nonplanar were removed using a plane fitting algorithm [41]. This helped in removing the support mechanism consisting of a small rod like structure that connects 3DUID to the wall. This produced an offset between the 3DUID and the wall in the point cloud for recognition. A connected component based segmentation was initially tested to separate 3DUIDs against the wall. However, the connected component based segmentation missed several 3DUID tags in the point cloud when the variation in the surface around 3DUID tags was large. Therefore, a new algorithm called 3DReG, which uses an ordered point cloud was developed for more robust recognition.

Locating and identifying patterns is challenging in an unordered point cloud due to uneven point densities and unknown relative positions between the structures (Dong et al., 2018). Therefore, the point cloud was first ordered using the "ordering of points to identify the clustering structure" (OPTICS) algorithm which helps in visualising the density based clustering structure of the point cloud [42]. In this study, a modified version of OPTICS algorithm using a perimeter of a triangle was used [43]. Clustering was performed on an ordered point cloud based on three handcrafted parameters between subsequent point including Euclidean distance, range difference and point angle. Range difference ($|R|$) is measured between the range of the first point and the second point, where the range is measured from the point to the nearest trajectory. Point angle ($\theta$) denotes the angle between a line joining the first point ($P_1$) and the second point ($P_2$), and a line joining the first point ($P_1$) and the nearest sensor location (S). Since 3DUID was located at an offset distance from the wall, the Euclidean distance between the transition point on the wall and the 3DUID should be large. Transition point refers to the point where a transition occurs from one structure to another when moving subsequently in the ordered point cloud. In an ordered point cloud, a uniform transition between points, shown with a colour scale, can be observed (Fig. 6a). Also, the range values of all the points lying on the 3DUID are deemed to vary considerably when compared to other points in the surrounding (Fig. 6b). In Fig. 6c, $\theta_1$ represents the point angle between transition point $P_{1'}$ on the wall and $P_{2'}$ on the 3DUID whereas $\theta_2$ represents the point angle between points $P_1$ and $P_2$ on the 3DUID. Geometrically, the angle $\theta_1$ will always be small in comparison to $\theta_2$ due to offset between 3DUID and wall. A multiple threshold criterion with $|ED| > 8$ cm, $|R| > 8$ cm and $A < 45°$ or $A > 135°$ was used to segment 3DUIDs from the point cloud.

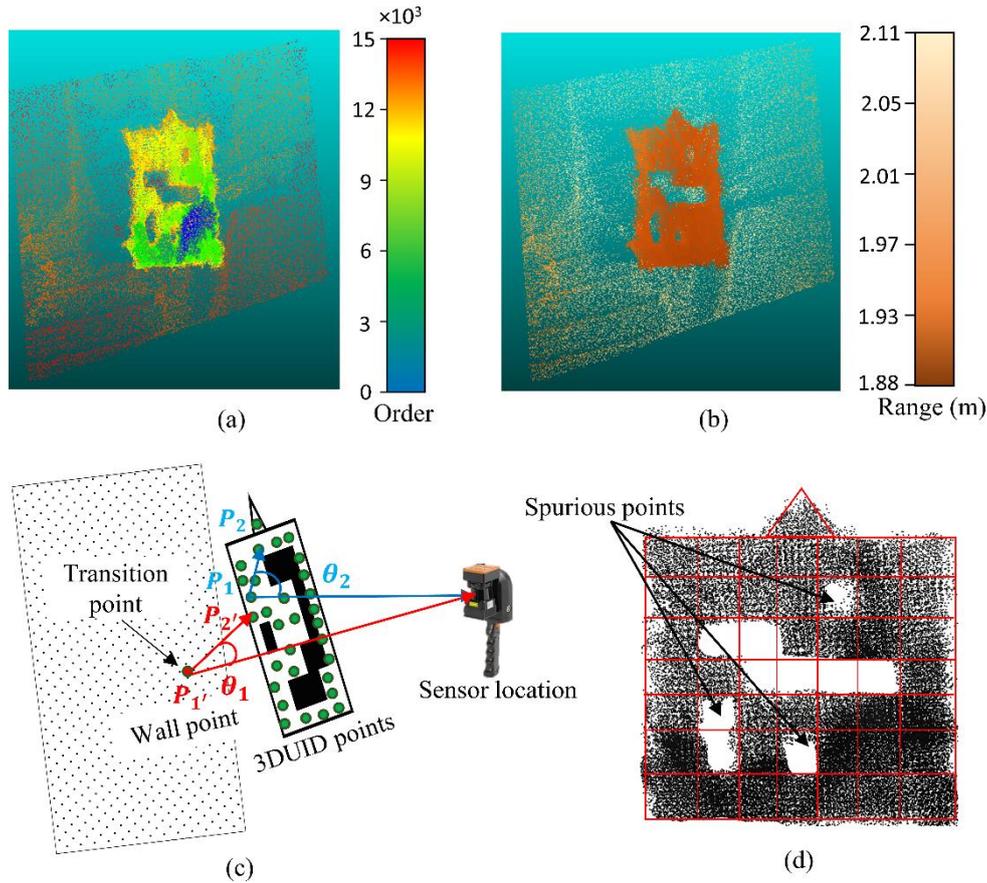

Fig. 6. (a) An ordered point cloud with point transitions shown using a colour scale, (b) point cloud coloured with respect to range showing the variability of 3DUID points with respect to surrounding points, (c) concept of point angle showing the angle of point transition with respect to the nearest sensor location. Transition point shown in red denotes that after this point transition from wall to 3DUID occur. (d) Some of the spurious points encountered in the void region of 3DUID due to sensor range inaccuracy.

In the clustered data, a best-fit rectangle was computed for each cluster that was verified against the known 3DUID dimensions to identify potential 3DUIDs. A tolerance of ± 3 cm in the length and width of the fitted rectangle was set to account for probable sensor range errors. Henceforth, the point cloud cluster of potential 3DUID was divided into 6 cm × 6 cm grids, where the grid with points represented '1' and void of points represented '0' (Fig. 6d). Due to the range uncertainty of the laser scanner some spurious points were encountered in the void region that affected pattern recognition (Fig. 6d). Applying a simple threshold, efficiently helped in overcoming this issue and the pattern was simply decoded through binary conversion. Whenever there were less than 20 points in a grid it was considered '0' or else '1'. Upon matching of binary pattern with the installed 3DUID patterns, the point cloud of 3DUID and the corresponding tags were stored. The topmost point in the recognised 3DUID, denoting the tip of the notch, was obtained using singular value decomposition [44] for automated georeferencing.

### 2.3.4   Georeferencing, coregistration and accuracy analysis

A rigid body translation and rotation of the scanned point cloud was carried out to establish a relationship between local and global coordinate systems utilising surveyed notch of 3DUID as GCT. The local coordinates of 3DUID's were obtained automatically using the recognition algorithm presented in Section 2.3.3 while global coordinates were obtained using the total station survey. Horn's quaternion-based algorithm was implemented to achieve a transformation matrix between the two coordinate systems [45]. In Horn's quaternion approach, the best translation offset is the difference between the centroid of the coordinates in the reference system and the centroid of the coordinates in the transformed system, i.e., rotated and scaled. The best scale coefficient is obtained through the ratio of root mean square deviations, of coordinates in two (reference and transformed) systems from their respective centroids. Additionally, the best rotation matrix is represented by the unit quaternion obtained from the eigenvector associated with the highest eigenvalue of a symmetric $4 \times 4$ matrix. The elements of the symmetric matrix are formed by the combinations of sums-of-products of corresponding coordinates of the points between reference and transformed systems [45]. The Horn's quaternion approach is preferred over other iterative empirical, graphical and numerical procedure as it is non-iterative, does not require an initial approximation and provides best possible transformation directly in a single step with statistical measures of points in two coordinate systems.

Assessment of error in georeferencing is typically measured at a set of check GCTs when registration is performed using a set of reference GCTs. A leave-one-out cross-validation (LOOCV) procedure was adopted to estimate the performance of georeferencing of point cloud scans using 3DUID tags, at four discrete spacing levels (Fig. 3f). In the process, the georeferencing error was calculated for each possible combination by selecting four 3DUID as reference GCTs to obtain a conversion matrix, and then determining error on the converted coordinate for the fifth 3DUID as check GCT. The mean absolute error (MAE) from all possible combination was used in the assessment of georeferencing accuracy. Additionally, a mutual cloud-to-cloud (C2C) distance assessment was performed for the two georeferenced point clouds scans collected from the same area. The C2C distance was measured for each of the four discrete spacing level using all of the five 3DUID's as reference GCTs.

Automated coregistration of point clouds without point picking is computationally intensive. A commonly practiced solution is to downsample the point cloud for performing a quick sparse-registration, followed by a subsequent fine-registration [46,47]. In this study, the automated data coregistration using 3DReG was achieved by first recognising 3DUID tags in two different point clouds using algorithm mentioned in Section 2.3.3, and then establishing a correspondence using Horn's quaternion based approach using coordinates of unique 3DUID tags. The two point clouds were then coarsely aligned considering one of the point clouds as reference. The roughly aligned point clouds were finely refined using a rigid coregistration through iterative closest point algorithm (ICP). The ICP algorithm iteratively minimises the root means square error calculated from Euclidean distances between points in the two point clouds [48]. The algorithm requires a rough initial alignment of point clouds for an accurate result which was provided using unique 3DUID tags. A non-rigid coregistration algorithm was avoided for fine refinement as it distorts the original/natural characteristics of the point cloud to match the reference point cloud, which is unsuitable for an application like change detection. The accuracy of coregistration was evaluated by calculating the median of C2C distances between the reference and transformed point clouds.

The output of 3DReG algorithm was compared against widely used coregistration algorithms ICP and normal distribution transforms (NDT) which were implemented independently without the aid of 3DUID in the process. In NDT, the point cloud is divided into voxel and then a normal distribution, which locally models the probability of measuring a point, is assigned to each voxel [7,49,50]. A Newton's algorithm is then used to match other scans using piecewise continuous and differential probability density. The NDT algorithm does not require initial alignment of point clouds, and the accuracy and processing time depends on the defined voxel size (0.5 m in this study). The two algorithms were chosen due to efficiency in processing time. An ICP algorithm was preferred over NDT in 3DReG processing workflow as the solution of NDT does not depend on initial point cloud alignment, while ICP exploits the similarity between the two point clouds using rough initial alignment for quick convergence.

### 2.3.5 Extraction of roadway clearance

Assessment and routine monitoring of roadway profile, mainly horizontal and vertical clearance, is essential to estimate constructional requirement for vehicle passage and measure changes, such as convergence and floor heave, resulting from mining activities. The automated monitoring of roadway profile has been a challenging task due to incorrect correspondences of multi-temporal data because of lack of reference points. The 3DUID tags play crucial role in providing reference and roadway profile between two 3DUID tags can be efficiently compared. In the experiment, the profile of a roadway between first 3DUID and last 3DUID was automatically extracted from the 3D point cloud and was compared against the ground truth measurement collected every 10 m using laser distometer (Leica Geosystems, St. Gallen, Switzerland).

## 3. Results

The two scanned point clouds, approximately 850 m in length, after SLAM processing contained a total of 37.6 million and 46.6 million points with a point density of 3156 and 3341 points/m$^2$, respectively. A slight difference in point density was observed due to minor variation in vehicle speed between two scans, affecting the total time taken to scan the given area, with a slower vehicle speed run increasing the point density. To assure that the two point clouds do not have mapping drifts, the distances between subsequent 3DUID tags in the two datasets were compared mutually, and verified against the ground truth distances obtained through a total station survey. The MAE was 0.14 m between the point clouds, 0.20 m between the first point cloud and total station survey, and 0.28 m between the second point cloud and total station survey. Since the two point cloud had a small MAE using 3DUID tags, it can be concluded that the mapping drift incurred was small. A relatively more MAE between a dataset and total station survey might be due to surveying errors.

### 3.1 3DUID recognition

The raw point cloud obtained from SLAM was pre-processed and segmented for 3DUID recognition using the automated approach presented in Section 2.3.3. The recognised 3DUID tags in the point cloud were uniquely colour coded to represent unique identities for visualisation purpose (Fig. 7). The offset present between the wall and installed 3DUID tags enabled pattern decoding using depth perception. The OPTICS based algorithm has high initial computation time for ordering point cloud through data triangulation which was substantially reduced by removing irrelevant roof and floor points using a strategic segmentation. Out of the thirteen 3DUID tags installed across the mine section, all of them were recognised using 3DReG giving a detection and recognition accuracy of 100 %. The employed pattern recognition technique was robust and performed well under challenging environmental factors such as sub-optimal light, limited surface reflectivity, surrounding features and dusty conditions. In contrast, the alternative connected component based segmentation missed five out of thirteen 3DUID tags and only provided a recognition accuracy of 61.54 %. Local variations on surfaces were present around 3DUID which affected connected component based segmentation (Fig. 7). Such erroneous variations were filtered in the ordered point cloud using multiple thresholds. The 3DUID tags were installed at the turning/intersection/crosscuts and along the longer section of the mine roadway.

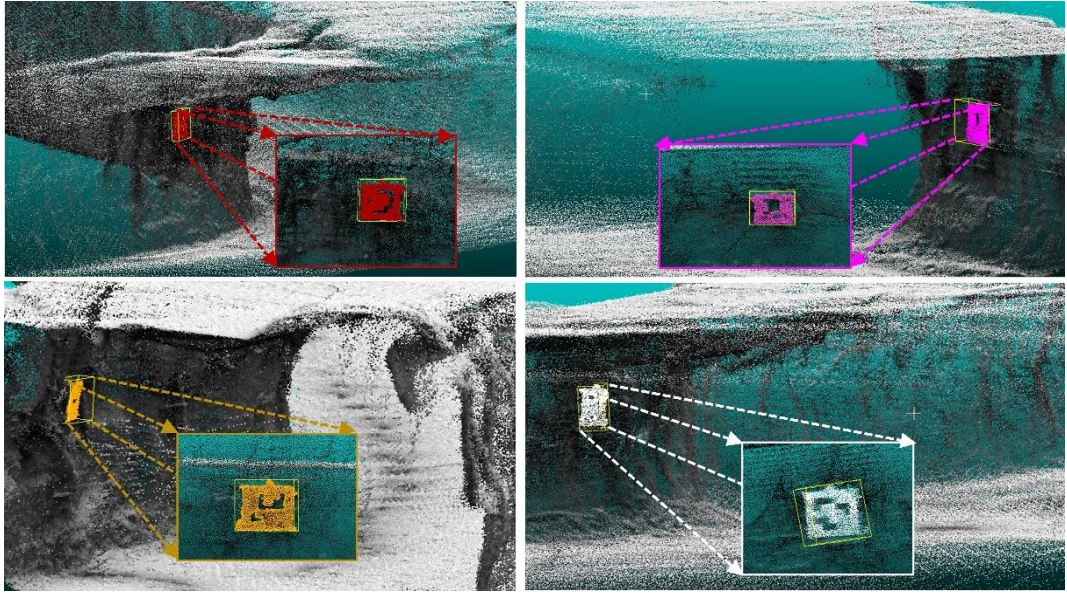

Fig. 7. Recognised 3DUID patterns are shown in yellow bounding boxes in the point cloud of an underground mine. The colour of 3DUID in the point cloud represents a unique identity tag associated with 3DUID.

## 3.2 Georeferencing

The georeferencing of sections with different 3DUID spacing was analysed with a LOOCV approach for each section. The mean and best case results obtained for various subsections are shown in Fig. 8a. The MAE represents the average of absolute errors achieved after evaluating five different combinations according to LOOCV analysis for five 3DUIDs (see Fig. 3f). A measure of best case absolute error was also provided to represents the most accurate georeferencing amongst the five possible combinations. A systematic increase in error was observed with an increase in spacing between 3DUID tags with a substantial increase beyond 100 m spacing (Fig. 8a). To further evaluate the georeferencing accuracy, a median of C2C distances between the two georeferenced datasets for subsections with different 3DUID spacing was computed. The C2C distance increased gradually with minor variations between 25 m and 100 m spacing levels, and followed a sharp increase beyond 100 m (Fig. 8b). Table 1 reflects quantitative estimates of errors for various spacings levels between 3DUIDs. The error in georeferencing was within 1 m when 3DUIDs were installed at a distance of 100 m or less. The georeferencing error increased substantially i.e. exceeding 2 m when spacing amongst 3DUIDs reached 200 m. Based on these results, nine 3DUID tags installed at 100 m apart were used for final georeferencing of the entire 850 m point cloud scan. The error in georeferencing measured by calculating MAE between the surveyed coordinates and the transformed point cloud coordinates was 1.76 m. The last 3DUID exhibited the highest error in the georeferenced coordinate, the absolute deviation was 1.83 m along x-axis, 0.82 m along y-axis and 0.014 m along z-axis. This could be due to the error accumulated in total station survey which increases with survey distance along the lengthwise direction of the tunnel i.e. the x-axis. The two point clouds in their independent local coordinate system are shown in Fig. 8c, while the two point clouds after georeferencing using 3DReG is shown in Fig. 8d. The median of C2C distances between the two point clouds after georeferencing was 0.5 m indicating a decent overlap between the two point cloud scans.

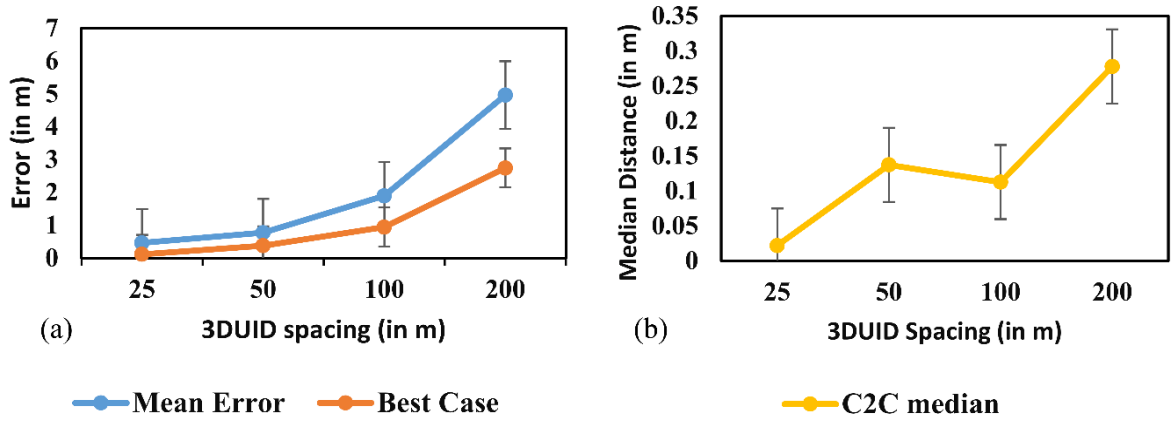

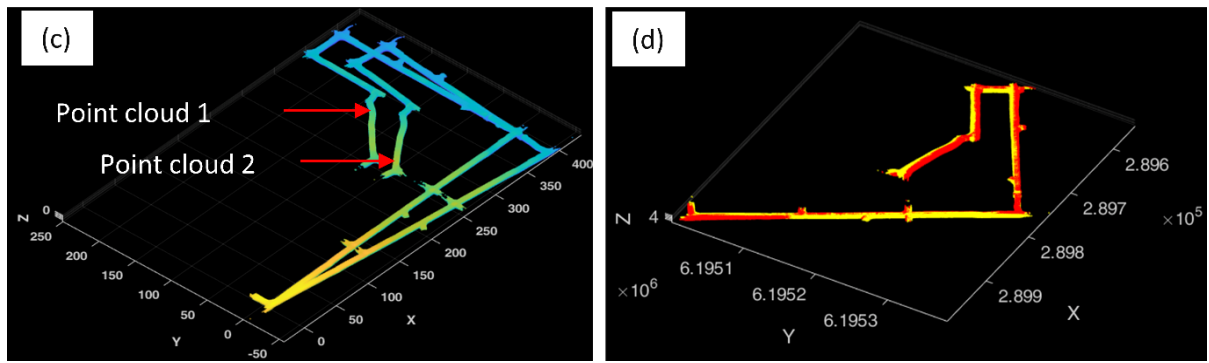

Fig. 8. (a) Error between georeferenced coordinate and corresponding surveyed coordinate through LOOCV error analysis for different 3DUID spacing, (b) a cloud to cloud distance comparison between two georeferenced point cloud achieved using 3DUID tags with varying spacing, (c) the two point cloud in local coordinate system achieved after SLAM processing, and (d) georeferenced/ globally aligned point cloud.

Table 1. Mean absolute error in georeferencing for sections with different 3DUID spacing levels.

| 3DUID spacing (in m) | MAE of LOOCV (in m) | Best case absolute error in LOOCV (in m) | Median of C2C distance between the two datasets (in m) |
|---|---|---|---|
| 25 | 0.46 | 0.11 | 0.02 |
| 50 | 0.78 | 0.38 | 0.14 |
| 100 | 1.89 | 0.94 | 0.11 |
| 200 | 4.96 | 2.74 | 0.28 |

### 3.3 Coregistration

In this study, three rigid coregistration methods, ICP, NDT and 3DReG were evaluated. The accuracy of the three approaches was evaluated by comparing the median of cloud to cloud (C2C) distances between the coregistered point clouds. The reference and transformed point cloud are shown in Fig. 8c. The NDT algorithm was not able to align point clouds and there was a substantial mismatch between the reference and the aligned point cloud (Fig. 9a). In contrast, the ICP algorithm had a better performance; however, a misalignment was observed at some cross-cuts of the mutually aligned point cloud (shown with a closeup view in Fig. 9b). The 3DReG algorithm, which uses 3DUID tags for coarse alignment and ICP for fine refinement, provided the best achievable result with perfect overlap between the two point clouds (Fig. 9c). The difference in the achieved coregistration can be observed visually through the closeup view of the cross cuts (Fig. 9b and 9c) and quantitatively through Table 2 which lists the median of C2C distances observed in two georeferenced point clouds and coregistered point clouds. The median of C2C distances for the 3DReG algorithm was 0.16 m which was a significant improvement over ICP and NDT coregistration algorithms that exhibited an error exceeding 0.5 m. Georeferencing of two point clouds using surveyed 3DUID tags had a second best performance in terms of the median of C2C distance that was 0.50 m.

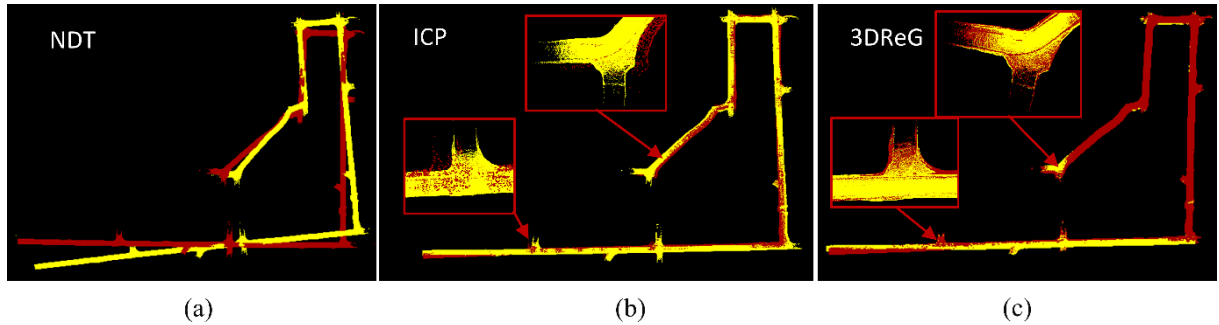

(a) (b) (c)

Fig. 9. (a) Point cloud coregistered using NDT algorithm. An inaccurate alignment of two point clouds can be observed. (b) Point clouds coregistered using the ICP algorithm. Mismatching can be seen at some cross-cuts in the mine section. (c) Point cloud coregistered using 3DReG where 3DUID is used for initial alignment followed by a fine refinement of the point cloud using ICP. A perfect overlap between two point clouds is visible.

Table 2. A cloud to cloud distance comparison between coregistered and georeferenced point cloud.

| Georeferencing and coregistration method | Median C2C distance (in m) | Time | Iterations | Sampling |
|---|---|---|---|---|
| Georeferencing using surveyed 3DUID | 0.50 | 5 sec | 0 | - |
| ICP | 0.69 | 17.4 sec | 100 | 1,000,000 |
| NDT | 6.93 | 32 minutes | 100 | 1,000,000 |
| 3DReG | 0.16 | 20 sec | 100 | 1,000,000 |

### 3.4 Roadway profile extraction

An assessment of consistency between cross-sectional profile extracted automatically from point cloud using 3DUID and in-situ ground truth measurement was performed. This was primarily to ascertain the efficacy of extracting reliable cross-sectional profiles using point cloud scans, which is often a compliance and operational requirement for mitigating roadways collision hazards to moving vehicles or mine personnel and measuring convergence or floor heave rate. The cross-sectional profile obtained from the point cloud was validated by comparing vertical and horizontal clearance between the first and the last 3DUID against the field measurements collected using a laser distometer at every 10 m. The point cloud and extracted cross-sections at 10 m spacing are shown in Fig. 10a and 10b, respectively. The values of vertical and horizontal clearance along with error observed in measurement from a point cloud, represented with corresponding profile and error bar plots at each measurement point, is shown in Fig. 10c. The blue plot with red dotted points denotes vertical clearance while orange plot with black dotted point indicates horizontal clearance. The horizontal clearance is usually more than the vertical clearance to allow passage of incoming and ongoing vehicle without impedance. When summarised, the point cloud exhibited a root mean square error of 0.048 m and 0.065 m in vertical and horizontal clearance, respectively. The minimum and maximum error in the measurement of vertical clearance from point cloud was 0.001 m and 0.120 m, while in horizontal clearance was 0.001 m and 0.150 m, respectively.

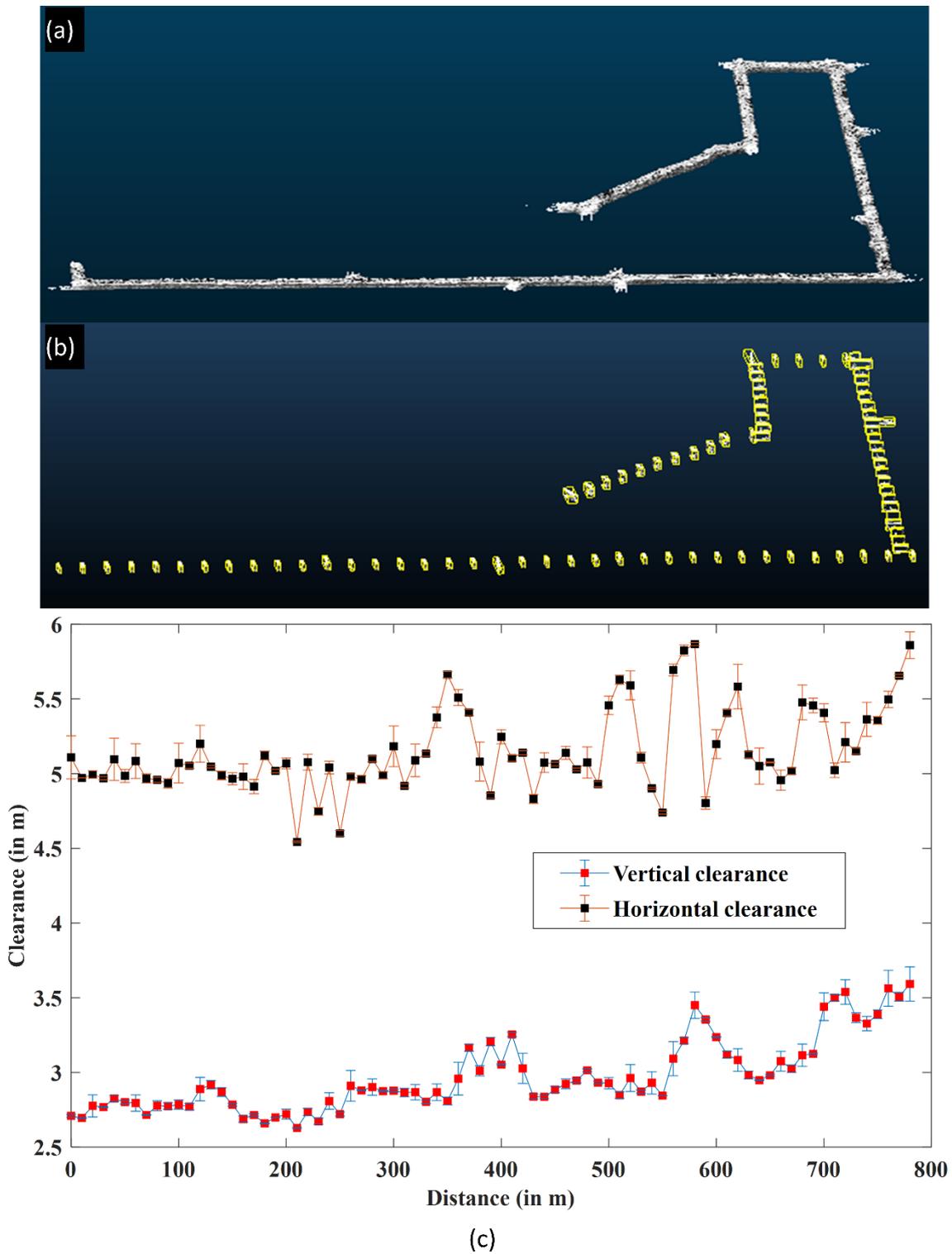

Fig. 10. (a) Point cloud of the test area, (b) cross-sections extracted at every 10 m starting from the first 3DUID and ending at last 3DUID, and (c) the observed error in actual vertical and horizontal clearance shown with box plot for the given length of the mine section.

## 4. Discussion

### 4.1 3DUID recognition

A set of initial laboratory tests were performed to obtain the minimum mappable unit of the laser scanner when a target is being scanned from varying distances, which was fundamental to the design of 3DUID. With an increase in scanning distance, the laser rays begin to diverge leading to uncertain points. The extent of beam divergence and in turn uncertainty depends on the maximum measurable range of the scanner. The initial laboratory tests provided a suitable benchmark of dimensions required for 3DUID development based on the characteristics of the mobile laser scanner used in this study. For instance, the ZebRevo user manual list the range accuracy of the scanner to be ± 3 cm for a scanning distance of 15 m – 20 m; however, the minimum mappable unit for a scanning distance of 5 m achieved through laboratory tests was still 3 cm. Further, the highly retroreflective material was found ineffective as it increased the noise because of specular reflection and led to many false points in the void region of 3DUID. The incurred noise affected the pattern decoding capability in the lab experiment.

### 4.2 Georeferencing and coregistration

The accuracy of georeferencing of a point cloud relied on the number of 3DUID reference tags and the spacing level. When 3DUID tags are placed closer (< 25 m) to each other, the MAE can be reduced to as low as 0.1 m. A systematic increase in the MAE was observed with an increase in spacing between the 3DUID tags as the local drift in scanning between the 3DUID tags is not captured. Therefore, based on the maximum permissible error limit for any particular application, the spacing between the tags should be decided. For instance, the change detection of a few centimetres in magnitude can be captured by placing 3DUID tags in close proximity. Conversely, for applications such as volumetric analysis where an error up to few metres are permissible, the 3DUID tags can be placed further apart.

The installation of 3DUID also facilitated the validation of georeferencing through the verification of horizontal and vertical clearance. In a GNSS denied and symmetrical environment, there is scarcity of reference points with respect to which the cross-sectional profile can be measured along the roadway/hallway. As such, it is challenging to match the cross-sectional profile in the two datasets that may provide crucial information on occurring changes in the sensitive environment. The 3DUID tags in such a case provide a reference and profile between the two 3DUID tags can be accurately and automatically matched. In this study, the matching of the cross-sectional profile validated the georeferencing where the minimum absolute error and maximum absolute error in the measured profile was 0.001 m and 0.15 m, respectively.

The coregistration of point clouds was addressed using the rigid transformation algorithms such as ICP and NDT instead of non-rigid algorithms as they do not distort the natural geometry of point cloud to match the collected scan with the reference scan. In the proposed 3DReG algorithm ICP was preferred over NDT because ICP utilises the initial alignment of two point clouds, obtained using 3DUID tags, to quickly converge to a better coregistration solution. Amongst all the evaluated coregistration algorithms, the NDT was found least accurate, as it determined the transformation matrix based on the voxels that exhibited similar probability density ultimately leading to a mismatch between the point cloud. Furthermore, the result of NDT depends on the selected voxel size and results in inaccuracy for larger voxel size. Therefore, the user while performing coregistration must be well aware of the point cloud resolution to select the optimal voxel size for better performance of NDT. The coregistration results of the ICP algorithm depend on the coarse initial alignment of point clouds and often produces false results in absence of it. Although the point clouds in this study had some initial alignment, the algorithm could not provide accurate coregistration due to structure similarity, repetitive features and lack of distinguishable points. Such conditions caused aliasing where false point matches led to minimisation of the distance between the point clouds. Therefore, a mismatch was observed at some of the crosscuts in the mine section. Aliasing issue was successfully resolved in 3DReG algorithm by accurate point to point matching using unique 3DUID tags. Unlike ICP and NDT, the 3DReG algorithm is unaffected by the presence of discriminative features in the environment and can provide reliable results even when the environment is symmetric, feature deficient or highly repetitive. The accuracy of 3DReG algorithm is dependent on the spacing and distribution of 3DUID tags and the inter 3DUID distance should be kept low, preferably within 100 m, to keep the MAE in coregistration within 1 m for a point cloud of length 850 m.

In GNSS denied as well as complex underground space, achieving a state-of-the-art performance for georeferencing and coregistration is challenging. It is evident from the results that a sufficient number of discriminative features are required in the environment for accurate coregistration using conventional algorithms. The variations in the local environment are usually captured by defining several descriptors such as point normal, curvature, fast point feature histogram, eigenvalue descriptor, histogram of normals, anisotropy, scale invariant feature transform and rotation invariant feature transform to uniquely identify a place. Previous studies have mostly exploited handcrafted descriptors mentioned above or used machine learning algorithms such as deep convolutional neural network on known objects that are likely to be encountered for accurate coregistration [51,52]. For the environment that possesses high structural symmetry and similarity, the use of hand-crafted descriptor does not benefit. Moreover, for the environment with no prior training data or a limited number of defined objects such as in underground mines, machine learning based coregistration cannot be achieved. An alternative option in absence of distinguishable features is to provide additional information either through the use of multi-sensors that may include laser scanners, radar, inertial measurement unit, odometer, optical camera and infrared camera [53,54] or through modification of environment itself by adding distinguishable features such as GCTs [32,55]. The former addresses the concerns to some extent, however the cost and power consumption increases. Further, the use of additional power is constrained in a sensitive environment like an underground mine due to fire related hazards. Therefore, it is necessary that accurate mapping be achieved without relying on complementary sensors for additional information. The latter "GCT based approach" is an attractive option as the targets once installed can help in accurate mapping using a limited number of sensors thereby reducing system cost. The 3DUID tags used as GCTs in this study successfully demonstrated relatively accurate coregistration with an error as low as 0.16 m. The entire process from 3DUID recognition to coregistration was automated and point clouds of up to 850 m lengths were coregistered within 20 seconds on a system with 32-Core Processor 3.69 GHz and 256 GB memory (AMD Ryzen Threadripper 3970X, California, USA). The conventional algorithms such as NDT and ICP, as well as those using local descriptors, have a high computation time which is proportional to selected hyperparameters such as the number of points, voxel size and dimensionality of descriptors. The descriptor generation, in particular, is one of the most time-consuming steps where the processing is done on individual points or selected key points in the point cloud that may take up to hours to process depending on the system configuration. Coregistration using 3DUID tags overcomes time constraint while resulting in the least MAE, as the point clouds of only 3DUID tags are used for initial coarse registration which geometrically aligns the point clouds within 3 seconds. Since the point cloud already gets aligned, the time required by ICP for fine refinement reduces considerably and can be achieved in minutes.

The spacing as well as installation pattern of tags in the environment should be decided based on the application. It was observed that installation of tags along the longer sections of the tunnel may not improve coregistration or georeferencing of point clouds, as linearly placed GCTs are unable to constrain the coregistration solution of a 3D point cloud. Therefore, GCTs need to be well distributed in the scanned environment, which can be achieved by installing 3DUID tags along the intersections and turns.

## 4.3 Future application of 3DUID

The main application of 3DUID is to achieve accurate georeferencing and data coregistration which are one of the most fundamental aspects in 3D mapping required for localisation [56], surface reconstruction [57], change detection [58] and deformation monitoring [59]. The 3DUID tags proposed in this study need to be surveyed only once after installation and then all the point clouds can be automatic georeferenced or coregistered, which enables multi-temporal change detection without having to manually register multiple datasets. In the absence of surveyed coordinates for 3DUIDs, the proposed solution is still effective, as one of the scans can be used as a reference to automatically align other scans using 3DUID tags. Further, with the distances between each 3DUID tags known, either through a survey or based on reference map, relative movements (such as roadway tunnel subsidence, floor heaving and fracturing) occurring in an area can be efficiently tracked by measuring relative movement in the 3DUID tags or using C2C changes and similar change detection algorithms. An alternate use case benefit may involve a set of 3DUID tags installed at a stable location to act as an anchor point, with respect to which movement of surrounding structures could be effectively mapped with robustness. Furthermore, roadway profiles often get deformed in tunnels and underground mines due to compressive stress in rock mass activated as a result of removal of excavated materials, leading to floor or roof heaving and closure between walls. Routine measurement and monitoring of deformation in the roadway profiles has been challenging. Current practices still often involve manual in-field measurement approaches or include in-situ movement loggers such as extensometers [60,61] and closure meters [62]. These approaches are limited to provide a spatially sparse sampling of rock mass movement patterns. Numerical modelling and rock mechanics are dedicated field of research built on such in-field observations [63,64]. Laser scanning has been able to provide complementary data to enrich convergence modelling through rock mass stress prediction using numerical models. However, the laser scanning methods often become tedious with terrestrial laser scanners requiring frequent repositioning, extended duration of scanning and processing time involved in manual registration of scanned point clouds. Mobile laser scanning together with improvement in SLAM has addressed field related challenges in data acquisition to a large extent, however, effective utilisation of point cloud scans has remained a challenge in terms of manual georeferencing and coregistration. The proposed 3DReG algorithm with infield installed 3DUIDs is promising to address these existing limitations. The approached is deemed useful in providing better laser scanning capabilities to aid in monitoring and rock mass failure predictions, which is critical in avoiding expensive downtimes, and potentially fatal or non-fatal injuries.

Data processing in GNSS denied space involves two major components: (1) Frame by frame laser scan coregistration for simultaneous localisation and mapping (SLAM), and (2) georeferencing and coregistration of multi-temporal data for change detection applications. The study addresses the later, however, the former can be achieved following the presented approach for 3DUID recognition. SLAM based on GCTs is a well-researched topic [65]. However, the focus has been more on algorithmic improvement rather than the improvement in GCTs itself for quick recognition. In past, GCTs were either too hard to decode, difficult to construct and lacked a unique identity or relied on an additional sensor such as an optical camera for decoding [30,31,33]. The 3DUID tags presented in this study are easy to construct, simple to decode and do not require additional sensor apart from the laser scanner for recognition. The 3DReG algorithm proposed can be integrated into the SLAM algorithm for achieving accurate mapping and localisation through accurate loop closure detection. The subsequent laser frame matching can be done quickly and 3DUID tags can be stored as nodes, which can be used for feature matching and loop closure detection.

Underground mine automation has recently acquired significant interests for developing continuous miners, excavators and roadways haul vehicles. For efficient excavations, these machineries need to know their precise location with respect to the global mine map. Development of parallel technologies in driverless car industry such as laser sensing and autonomous navigation systems [66,67] is expected to benefit mining with the adoption of these technologies in underground spaces, towards a safer mining operation. The 3DUID tags can play a pivotal role in such cases where the accurate 3D georeferenced map obtained from onboard laser scanners sensors will help in localising the mine machinery or autonomous hauling vehicles.

## 5. Conclusion

Mapping of underground and indoor environments still poses specific challenges due to unavailability of a sensor positioning framework (GNSS), complicated structurally symmetric layouts, complex interactions between objects or repetitive features and occlusions. Geometrically accurate laser scanning assist in modelling fundamental infrastructure for applications in civil, mining and transportation industries. The presented study was aimed at overcoming practical challenges in seamless registration of point cloud scans collected in a complex underground mining tunnel. The solution involved employing a set of 3DUID tags used as unique and automatically recognisable GCTs to achieve reliable georeferencing and coregistration accuracy. The developed 3DUID tags are simple to design, construct, easy to install, does not require power to function and intrinsically safe for adoption in safety sensitive mining environments. A dedicated 3DUID recognition algorithm was developed to reliably extract and decode the 3DUIDs from the point cloud, with a hundred percent success rate. Furthermore, a simple and robust 3DReG algorithm was also designed to accurately georeference and coregister point cloud scans. Application specific utilisation of laser scanning could be streamlined by incorporating 3DUIDs and 3DReG algorithm in data collection and pre-processing pipelines, respectively. In this study, the application focused on mining examples of streamlined use case scenarios including modelling rock mechanics behaviour to predict structural failures, which provide an effective means to routinely monitor the condition of underground tunnels, and favour development of autonomous excavators and road haul vehicles. However, the demonstrated 3DUID technology could be used across other sectors requiring seamless mapping and reconstruction of the built environment. The utilisation of 3DUIDs is not limited to active laser scanning systems. In contrast, passive imaging systems such as optical, multispectral, hyperspectral and thermal sensors could be equally used. Potential research involving 3DUIDs remains open with opportunities in multi-modal sensor fusion and direct scene registration in SLAM.

**Funding**

This work was supported by the Australian Coal Industry's Research Program (ACARP) [Project number: C27057].